\documentclass[11pt]{article}
\usepackage{coling2016}
\usepackage{times}
\usepackage{url}
\usepackage{latexsym}
\usepackage[usenames, dvipsnames]{color}

\usepackage{multicol}
\usepackage{caption}
\usepackage{booktabs}
\usepackage{array}
\usepackage{amsmath}
\usepackage{amssymb}

\usepackage{graphicx}
\usepackage[utf8]{inputenc}

\usepackage{multirow}

\usepackage{cancel}
\usepackage[normalem]{ulem}

\newcommand\ie[1]{\textit{i.e.~}}
\newcommand\eg[1]{\textit{e.g.~}}

\title{Factored Neural Machine Translation }
  
  \author{\bf Mercedes Garc\'ia-Mart\'inez, Lo\"ic Barrault and Fethi Bougares\\
  LIUM, University of Le Mans\\
    {\tt FirstName.LastName@lium.univ-lemans.fr} \\}
  
\date{}

\begin{document}

\maketitle
\begin{abstract}
We present a new approach for neural machine translation (NMT) using the morphological and grammatical decomposition of the words (factors) in the output side of the neural network. 
This architecture addresses two main problems occurring in MT, namely dealing with a large target language vocabulary and the out of vocabulary (OOV) words.
By the means of factors, we are able to handle larger vocabulary and reduce the training time (for systems with equivalent target language vocabulary size). 
In addition, we can produce new words that are not in the vocabulary.
We use a morphological analyser to get a factored representation of each word (lemmas, Part of Speech tag, tense, person, gender and number). 
We have extended the NMT approach with attention mechanism~\cite{Bahdanau} in order to have two different outputs 
%(originally available with only one output corresponding to words)
, one for the lemmas and the other for the rest of the factors.
The final translation is built using some \textit{a priori} linguistic information. 
We compare our extension with a word-based NMT system. 
The experiments, performed on the IWSLT'15 dataset translating from English to French, show that while the performance do not always increase, the system can manage a much larger vocabulary and consistently reduce the OOV rate. 
We observe up to 2\% BLEU point improvement in a simulated out of domain translation setup.
%We modified the standard NMT architecture to enable the generation of two different streams of symbols: the lemmas and the factors.
%Those factors are used to train only with these outputs in the target language instead of training with the full vocabulary with the completed words. 
%This idea could perform better in morphologically richer languages like French compared to English. 
\end{abstract}

\section{Introduction}
\label{intro}

%!TEX root = fnmt.tex

Neural Machine Translation (NMT) has been further developed in the last years~\cite{Bahdanau}. 
In contrast to the traditional phrased-based statistical machine translation~\cite{Koehn} that automatically translates subparts of the sentences, NMT uses the sequence to sequence of words approach~\cite{Cho}. 

Recently, NMT has improved the results of the phrased-based systems~\cite{Bahdanau}.
Besides these improvements in NMT, some problems still remain. 
One problem is the high computational cost of the target word probability due to the softmax that requires to normalize all the output values, see Equation~\ref{eq:softmax}:
	\begin{equation} \label{eq:softmax}
	p_i =  e^{o_i} / \sum_{r=1}^N e^{o_r} ~\mathrm{for}~ i \in \{1, \dots, N\}
	\end{equation}
where $o_i$ are the outputs, $p_i$ their softmax normalization and $N$ the total number of outputs.
%Then, the vocabulary size of the target words is limited in NMT, it cannot be very large. 

In order to solve this issue, a standard technique is to define a short-list containing the most frequent words only.
This has the disadvantage of increasing the out of vocabulary (OOV) rate. 
OOV words correspond to those unseen in the training dataset or which are not included in the vocabulary.
%All the words not included in the short-list are considered as unknown (UNK) words. 
They are all considered as unknown words and mapped to the special UNK token.

%Other works propose various techniques to solve this problem.
\newcite{Jean}, proposed to carefully organise the batches so that only a subset of the target vocabulary is possibly generated at training time. 
This allows the system to perform the softmax only on this subset during training (the complexity remains the same at test time).
Another possibility is to define a structured output layer (SOUL) to handle the words not appearing in the shortlist. 
This allows the system to always apply the softmax normalization on a layer with reduced size \cite{Le}.

Recently, some works have used subword units to translate instead of words.
In \newcite{Sennrich}, the rare and some unknown words are encoded as subword units with the Byte Pair Encoding (BPE) method. 
The authors show that this can also generate words which are unseen at training time.
As an extreme case, the character-level neural machine translation has been presented in several works~\cite{Chung,Ling15,Costa-Jussa} and showed very promising results.

In this work we propose an approach using factors as unit level in the output side of the neural network. 

The factors are referring to the linguistic annotation at word level like the Part of Speech (POS) tags. 
Moses toolkit~\cite{Haddow} for statistical machine translation is able to manage factors information in addition to the words to be able to improve the translation. 
Some works have used factors as additional information for language modeling~\cite{Bilmes,Alexandrescu}.
Recently, factors have been used as linguistic input features to improve NMT~\cite{SennrichH16} as well.

Our approach differs from previous works in the sense that we use only the linguistic decomposition of the words in the output side.
Each word is represented by its lemma along its linguistic factors (POS tag, tense, gender, number and person).
By these means, the target vocabulary size is reduced because we do not have to keep all the derived forms of the verbs, nouns, adjectives, etc. 
Furthermore, we are able to produce new words that are not in the vocabulary using all the derived forms of the lemmas. 

%As described in Figure~\ref{factors_schema}, 
We use two different outputs for the translation at word level, one output is the lemma of the word and the other output is the rest of the factors mentioned earlier.
% that are corresponding to the POS tags, tense, gender, person and number of the word. 
Multiple output neural networks have been used before \cite{FiratCB16} with the difference that in our approach the system produces both outputs at the same time instead of scheduling them. 
With both outputs (lemma and factors) we are able to generate the final word using linguistic resources. 
%\begin{figure}[!htbp]
%\centerline{\includegraphics[scale=0.25]{2outputs}}
%\caption{\label{factors_schema} Factors schema approach for NMT}
%\end{figure}
%We produced some experiments using English as source language and French as target language, which is morphologically richer than English.

%
% The following footnote without marker is needed for the camera-ready
% version of the paper.
% Comment out the instructions (first text) and uncomment the 8 lines
% under "final paper" for your variant of English.
% 
\blfootnote{
    %
    % for review submission
    %
%    \hspace{-0.65cm}  % space normally used by the marker
%    Place licence statement here for the camera-ready version, see
%    Section~\ref{licence} of the instructions for preparing a
%    manuscript.
%    %
    % % final paper: en-uk version (to license, a licence)
    %
  	\hspace{-0.65cm}  % space normally used by the marker
         This work is licensed under a Creative Commons 
         Attribution 4.0 International Licence.
         Licence details:
         \url{http://creativecommons.org/licenses/by/4.0/}
    % 
    % % final paper: en-us version (to licence, a license)
    %
    % \hspace{-0.65cm}  % space normally used by the marker
    % This work is licenced under a Creative Commons 
    % Attribution 4.0 International License.
    % License details:
    % \url{http://creativecommons.org/licenses/by/4.0/}
}

\section{Neural Machine Translation}
\label{nmt}
%!TEX root = fnmt.tex

%Neural Machine Translation with attention mechanism \cite{Bahdanau2014}

% Explanation of RNN encoder-decoder
The encoder-decoder architecture used for NMT consists of two recurrent neural networks (RNN), one for the encoder and the other for the decoder. The encoder maps a source sequence into a continuous space representation and the decoder maps the representation back to a target sequence.
%At each time step, the hidden state of the RNN is updated.
%the hidden state is controlled by a sophisticated units in order to memorize information or delete it.
Our trained neural translation models are based on a bidirectional encoder-decoder deep neural network equipped with an attention mechanism~\cite{Bahdanau}, as described in Figure~\ref{nmt}.

\begin{figure}[!htbp]
\centerline{\includegraphics[scale=0.25]{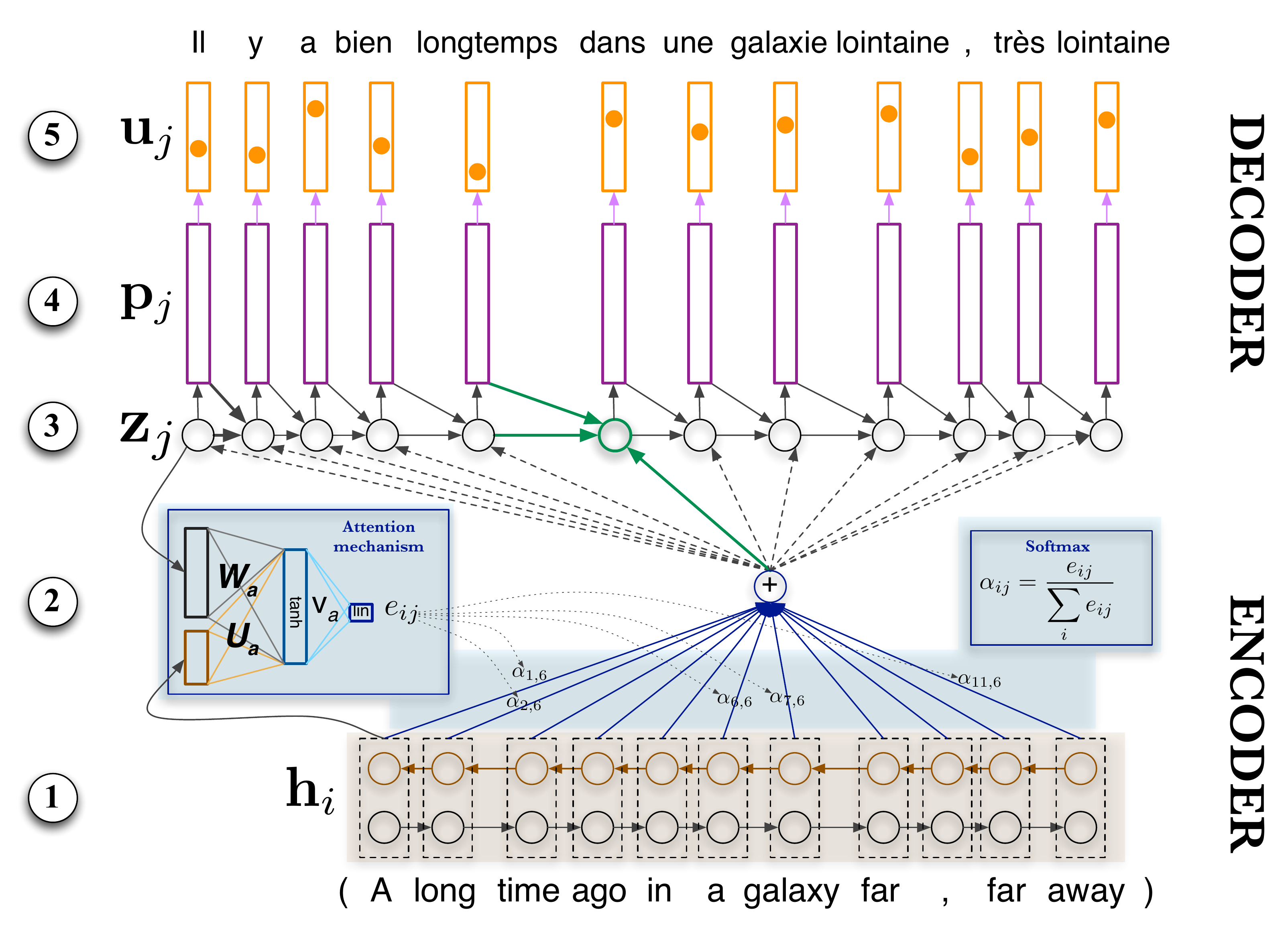}}
\caption{\label{nmt_model} Architecture of the NMT system equipped with an attention mechanism.}
\end{figure}

This architecture consists of a bidirectional RNN as an encoder (as seen in stage 1 of Figure~\ref{nmt}).
An input sentence is encoded in a sequence of annotations (one for each input word), corresponding to the concatenation of the outputs of a forward and a backward RNN. 
Each annotation represents the full sentence with a strong focus on the current word. 
The decoder is composed of a conditional RNN as provided for the DL4MT winter school\footnote{https://github.com/nyu-dl/dl4mt-tutorial} (see stage 3 of Figure~\ref{nmt}), equipped with an attention mechanism (stage 2).
%\vspace{-1.5cm}
The attention mechanism aims at providing weights for each annotation in order to generate a context vector (by performing a weighted sum over the annotations). The attention mechanism uses the hidden state at timestep $j$ of the decoder RNN along with the annotation  $h_i$ to generate a coefficient $e_{ij}$. A softmax operation is performed over those coefficients to generate the annotation weights $\alpha_{ij}$.
As described in \cite{Bahdanau}, the annotation weights can be used to align the input words to the output words.
The RNN takes as input the context vector, the embedding of the previous output word (stage 4), and of course its hidden state. 
Finally, on stage 5 of the Figure~\ref{nmt}, the output probabilities of the target vocabulary are computed. 
The word with the highest probability is selected to be the translation at each timestep.
The encoder and the decoder are trained jointly to maximize the conditional probability of the correct translation.

%\vspace{-3.5cm}
\section{Factors in Neural Machine Translation}
\label{factors}
%!TEX root = fnmt.tex

%\vspace{-1.5cm} 
%\vspace{-20pt} 
\begin{tabular}{p{10cm}@{\hskip 25pt} p{4.5cm}}
To perform factored neural machine translation, we need to extend the standard NMT architecture of the Figure~\ref{nmt} to allow generating several output symbols at the same time.
For the sake of simplicity, we decided to generate only two symbols: the lemma and the concatenation of the different factors that we considered.
%wo=devient po=v le=devenir te=P pe=3 ge=# nu=s ca=# ty=#
For example, from the word \emph{devient}, we obtain the lemma \emph{devenir} and the factors \emph{vP3\#s} meaning that it is a \textbf{v}erb, in \textbf{P}resent, \textbf{3}rd person, irrelevant gender (\textbf{\#}) and \textbf{s}ingular.
The morphological and grammatical analysis is performed with the MACAON toolkit \cite{macaon}.
%\footnote{MACAON website: http://macaon.lif.univ-mrs.fr/index.php?page=home-en} 
Figure~\ref{fi:factors_architecture} shows this modification. & 
    \vspace{-8pt}
    \includegraphics[scale=0.3]{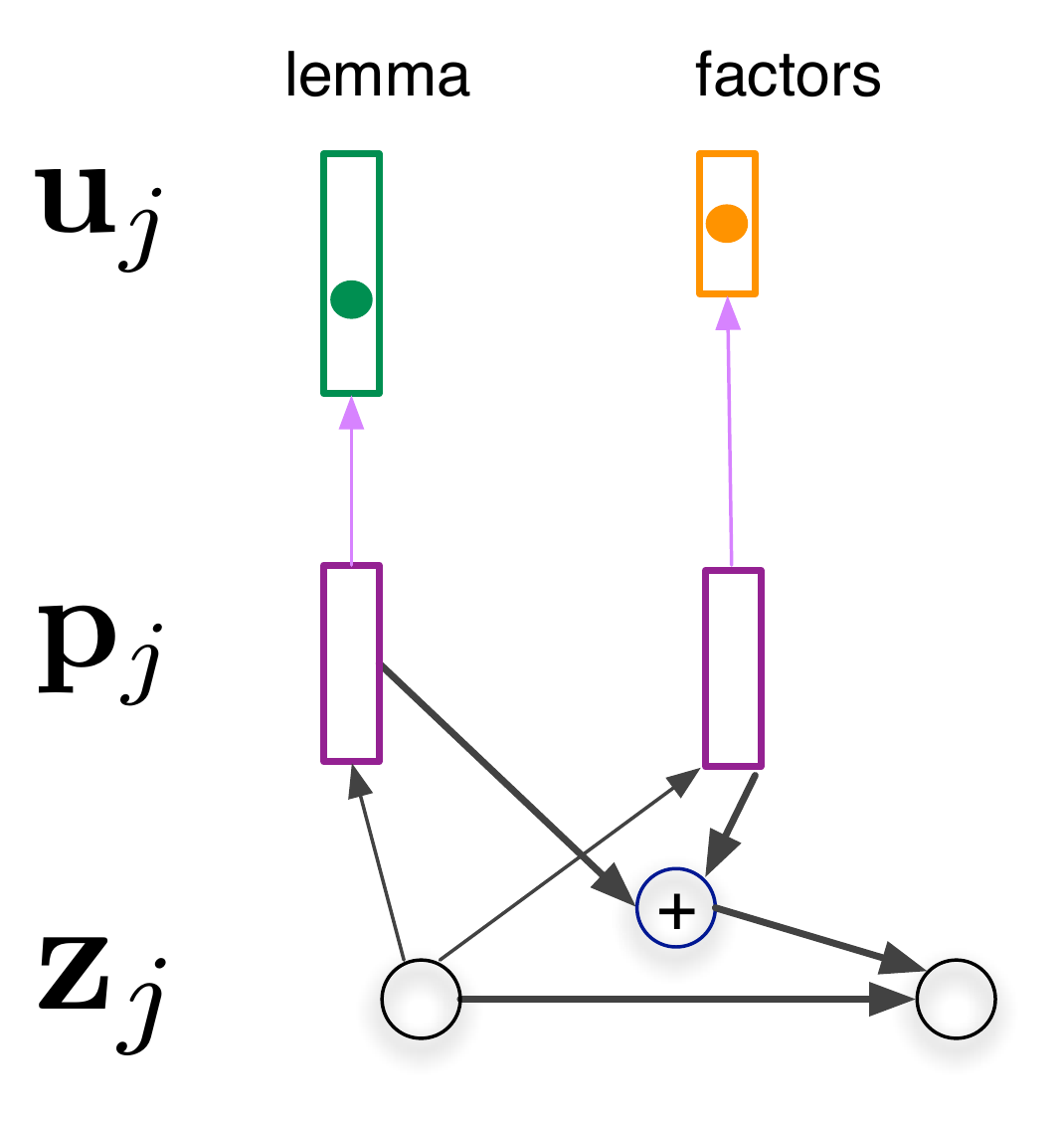}
    \captionof{figure}{\label{fi:factors_architecture} Output detail of the Factored NMT architecture.}
\end{tabular}

% ?mention cost = cost (lemma) + cost (factors)?
%We use as cost of the neural network the sum of the cost of the lemmas and the factors
%\vspace{-0.5cm} 
As we can see, lemmas and factors are generated separately, which in some cases, lead to sequences with different length.
To solve this problem, we give priority to the length of the lemmas. 
Consequently, we constraint the length of the factors sequence to be equal to the length of the lemma sequence. 
This is motivated by the fact that the lemmas are closer to the final objective (a sequence of \textbf{words}) and that they are the symbols carrying most of the meaning. 
%cutting the hypothesis of the factors until the length of the lemmas and always generating output of factors until to reach the length of the lemma sequence.

Another issue is the feedback that the RNN receives. 
In the word based model, this feedback is the embedding of the previous generated word. 
Since we have two outputs, we have to decide what will be given to the decoder RNN.
Several options are possible and will be explored in this paper (details in section~\ref{sec:feedback}).
For the first set of experiments, only the previous lemma embedding was used as feedback (no information from the factors output).

\subsection{Handling beam search with factors}
%\begin{tabular}{p{10cm}@{\hskip 35pt} p{3.5cm}}
%    \vspace{10pt} 
The beam search procedure has also been extended with respect to the original approach, since we are actually facing two beams (one for lemmas and one for factors).
We need to deal with the multiple outputs because we do not want to rely solely on the lemma sequence to decide which are the best sequences.
Then, we merge the two beams. % as shown in Figure~\ref{fi:beamsearch}. 
Once the best lemma and factors hypotheses are generated for each partial hypothesis, the cross product of those output spaces is performed. 
By this mean, each lemma hypothesis is associated with each factors hypothesis.
Afterwards, we keep the $k$-best combinations for each sample, with $k$ being the beam size.
Finally, the number of best hypotheses is reduced again to the beam size for further processing.
%&
%%\begin{figure}[!htbp]
%    \vspace{0pt}
%\centerline{\includegraphics[scale=0.20]{beamsearch}}
%\captionof{figure}{\label{fi:beamsearch} Beam search with factors (beam size = 3).}
%\end{tabular}
%\end{figure}

% Explanation of how we get the factors (MACAON) 
% /lium/buster1/garcia/workspace/script/run\_factors.csh
%MACAON POS analyser\footnote{MACAON website: http://macaon.lif.univ-mrs.fr/index.php?page=home-en} has been used to get the factors (lemma, POS tag, tense, person, gender and number) from the dataset. 
%MACAON is a tool suite for Natural Language Processing developed specially for French and English languages.

% Explanation factors2word 
\subsection{From factors to word}
Once we obtain the factorized outputs from the neural network, we need to fall back to the word representation.
This operation is also performed with the MACAON tool, which, given a lemma and some factors, provides the word candidate. %a list of word candidates.
%we make use of a dictionary with 656K entries of words with their corresponding factors. 
%This dictionary has been also produced with the MACAON toolkit. 
\section{Experiments}
\label{experiments}
%!TEX root = fnmt.tex

\newcommand{\ig}{i}
\newcommand{\ug}{z}
\newcommand{\rg}{r}
\newcommand{\hg}{h}

\newcommand{\mat}[1]{#1}
\newcommand{\scu}[1]{\sc{\underline{#1}}}
\newcommand{\vect}[1]{#1}
\newcommand{\uvect}[1]{\underline{#1}}

\newcommand{\E}{E}
\newcommand{\U}{U}
\newcommand{\W}{W}

We performed several sets of experiments trying different architectures and vocabulary sizes for Factored NMT (FNMT) and comparing them with the NMT system.

\subsection{Data processing and selection}

We evaluate our approach on the English to French Spoken Language Translation task from IWSLT 2015 evaluation campaign\footnote{IWSLT'15: https://sites.google.com/site/iwsltevaluation2015}. 
%We use the parallel corpora provided by the organisers.
A selection method~\cite{Rousseau} has been applied using the available parallel corpora (news-commentary, united-nations, europarl, wikipedia, and two crawled corpora) and Technology Entertainment Design (TED\footnote{TED: https://www.ted.com}) corpus as in-domain corpus. 
We also do a preprocessing to convert html entities and filter out the sentences with more than 50 words for both source and target languages.
We finally end with a selected corpus of 2M sentences, 147K unique words for English side and 266K unique words for French side. 

\subsection{Training}
%train the models during several days.

We chose the following hyperparameters to train the systems. 
The embedding and recurrent layers have a dimensionality of 620 and 1000 respectively.
We use a minibatch size of 80 sentences trained with Adadelta algorithm. 
The norm of the gradient is clipped to be no more than 1~\cite{Pascanu} and the weights are initialized with \emph{Xavier}~\cite{Glorot}.
The validations start at the second epoch and are performed every 5000 updates. 
Early stopping is based on BLEU  with a patience set to 10 (early stopping occurs after 10 evaluations without improvement in BLEU).
The vocabulary size of the source languages is set to 30K. 
We varied the output layer size from 5K to 30K in order to simulate different levels of out of domain data. 
%For the factors approach, we have 30K vocabulary size for the lemmas and 142 for the factors.
%This allows to possibly generate 172K words with the MACAON tool. 
Once the model is trained, we set the beam size to 12 (as this is the standard value for NMT, \cite{Bahdanau}) when translating the development corpus.
%Investigations on the beam size should be performed in the future.

\subsection{Factors models and results}

The Factored NMT system aims at integrating linguistic knowledge into the decoder in order to obtain better performance when facing out of domain data and/or a low resource setup.
To assess the feasibility and estimate the potential gain of our approach, we performed a set of experiments reducing the output vocabulary size, simulating such an environment.
The results are presented in Table~\ref{ta:factors_res}.
\begin{table}[!htbp]
\begin{center}
\begin{tabular}{ l|c|c|c|c|c|c|c|c|c } 
 \hline
     &\multicolumn{2}{|c|}{\textbf {Output}} & & & & \multicolumn{3}{|c|}{\textbf {\%BLEU}} & \textbf{Oracle}\\
   \textbf{Model} & \textbf{size} & \textbf {vocab.} & \textbf{Coverage (\%)} & \textbf{\#OOV} & \textbf{\#Par.} & \textbf{word} & \textbf{lem.} & \textbf{fact.} &\textbf{word} \\
   \hline
   NMT 	& 30K 		& 30K 	& 97.96 		& 1775 	& 89.7M 	& 34.88 & - & - & - \\ %49K sec/epoch
   FNMT & 30K+142 & 172K	& 99.23 (+1.27) & 784 	& 89.8M 	& 34.80 & 37.78 & 42.72 & 36.33 \\ %51K sec/epoch
   \hline
   NMT 	& 20K 		& 20K 	& 97.03 		& 2171 	& 77.3M 	& 34.21 & - & - & - \\ %48K sec/epoch
   FNMT & 20K+142 & 139K 	& 98.88 (+1.85) & 1014 	& 77.4M 	& 34.46 & 37.52 & 42.65 & 36.14 \\ %time 49K sec/epoch
   \hline
   NMT & 10K 		& 10K 	& 94.51 		& 3996 	& 64.9M 	& 32.61 & - & - & - \\ %45K sec/epoch
   FNMT & 10K+142 & 85K 	& 97.72 (+3.21)	& 1897 	& 64.9M 	& 34.13 & 37.07 & 42.75 & 35.72 \\ %time 45K sec/epoch
   \hline
   NMT & 5K 		& 5K 	& 91.02 		& 6545 	& 58.7M 	& 30.54 & - & - & - \\ %time 44K sec/epoch
   FNMT & 5K+142 	& 48K 	& 95.61 (+4.59) & 3424 	& 58.7M 	& 32.55 & 35.22 & 42.98 & 33.86 \\ %time 45K sec/epoch
   \hline  
 \end{tabular}
\caption{\label{ta:factors_res} Comparison of the performance of the NMT and FNMT systems in terms of \%BLEU score evaluating at word level, and separately, each output lemma and factors. The size of the output layer and the size of the corresponding vocabulary are presented in columns 2 and 3. Columns 4 and 5 show coverage in test dataset and number of OOVs, respectively. Last column corresponds to the oracle output.}
\end{center}
\end{table}

The FNMT system obtains a similar performance compared to the NMT system (first two rows) in terms of word level BLEU score, despite the increased complexity of the architecture of our model (and in particular the two outputs). % and the linguistic resources. 

%potentiality
In order to estimate the capacity of such a model, we computed the oracle which corresponds to ignore the errors caused by the factors, \ie\ if we produce the correct lemma, then the correct word is generated (see last column of Table~\ref{ta:factors_res}).
We can see that a potential gain of more than 1.5\% BLEU points can be achieved with a perfect modeling of the factors, which is encouraging. % since this means that we can learn a more complex structure which can be beneficial if it can be trained efficiently.

% expressiveness
The first comment is that the Factored NMT approach is able to model a bigger word vocabulary while preserving manageable output layers size. 
This is due to the fact that the factors-to-word tool is able to generate words which are unseen in the training corpus, augmenting the expressiveness of our model.
For the sake of comparison, we provide the target vocabulary size for the standard NMT and the FNMT systems. 
For example, with an output layer size of 30K, the NMT system can model 30K words against 172K words for the FNMT system. 
This is an almost 6 times larger word vocabulary.

One consequence is that the word coverage is higher for the FNMT than for the NMT system, as shown in column 4.
However, for the first two systems (first two rows), we see that the difference between the coverages is small.
When decreasing the output layer size, we can observe that the coverage decreases slowly for FNMT systems compared to the word based system.
The FNMT approach surpasses standard NMT when the coverage difference becomes higher.
This proves that the approach is sound and well performing, when dealing with out-of-domain data. 
This is of course dependent on the linguistic knowledge available in the factors-to-word tool.
This is exactly the sought behavior: by integrating \textit{a priori} linguistic knowledge, we reduce the impact of the training conditions (domain, data availability, etc.) on the performance of the system.

%OOVs
The reduction of the out of vocabulary (OOV) rate of about 47\% is a promising result which is not always well reflected by the BLEU score.
These results would be better highlighted if performing a human evaluation (this point will not be addressed further in this paper).
To make things clear, the OOV rate corresponds to the number of UNK tokens generated by our system.
In those experiments, we did not use any specific method to replace them (\eg\ put source words aligned to them, use a dictionary, etc.)

%complexity: #params, ?training time?
Moreover, the number of parameters to train also decreases according to the size of the output layer, as shown in column 6, allowing a simpler training because we have to learn less weights in the model. 
For example, using a lemma output layer size of 10K instead of 30K (3 times smaller) for factored model, we obtain a small drop of 0.67 points in BLEU. %comparing FNMT with 10K and 30K output lemma size. 
By contrast, in NMT base model we observe a drop of 2.27 points in BLEU comparing the same output sizes 30K and 10K.

Another interesting remark is that the scores evaluating in lemmas and factors are higher than the BLEU in words for both systems, this is due to the difficulty of the final step to generate the words.
Nevertheless, the BLEU for factors are pretty low considering that the output layer size for this is only 142.
This can be due to two different causes.
First, the neural network is not able to correctly model this small output.
Second, the task of translating from English words to French factors is complex.

\subsubsection{Evaluating each output}

We evaluated BLEU at different levels (word, lemma or factors) using the base NMT system with only one output (see Table~\ref{ta:change_target_text}). 
We compare the values with the Factored NMT system results which models lemmas and factors at the same time. 
We observe that the difference between the results in BLEU for lemmas using the FNMT are similar to the NMT system. 
However, the differences evaluating factors are big between the two systems (2.44 difference of \%BLEU).
%\todo{and then what is your explanation? your analysis?}
This experiment confirms that the task to predict factors managing very different output sizes respect to the source words is not easy.
In future we will implement factors also in the input side of the neural network to verify this hypothesis. 
Also, we have to take into account that we are giving more priority to the length of the lemmas sequence than the factors one during beam search.
This also suggests that we adapt our architecture so that factors are better predicted to obtain a final better BLEU at word evaluation. 
\begin{table}[!htbp]
\begin{center}
\begin{tabular}{ l|l|c|c } 
 \hline
  &\multicolumn{3}{|c}{\textbf {\%BLEU}} \\
  \textbf{Model} & \textbf{word} & \textbf{lemma} & \textbf{factors} \\
  \hline
NMT 	& 34.88 & 37.72 & 45.16 \\
FNMT 	& 34.80	& 37.78 & 42.72 \\
 \hline
\end{tabular}
\end{center}
\caption{\label{ta:change_target_text}Comparison of the performances between standard NMT system and the Factored NMT system in terms of \%BLEU computed at word, lemma and factors level. The first line corresponds to 3 standard NMT systems built to generate at the output words, lemmas and factors, respectively.}
\end{table}

%%%%%%%%%%%%%%%%%%%%%%%%%%%%%%% cost weighting model %%%%%%%%%%%%%%%%%%%%%%%

%\input{factors_prediction}

%%%%%%%%%%%%%%%%%%%%%%%%%%%%%%% Feedback combination %%%%%%%%%%%%%%%%%%%%%%%%%%%%%%%%%%%%%%

%In order to better understand the behavior of the factored model, we will explore different architectures to improve it.

\subsubsection{Feedback}
\label{sec:feedback}
As explained in section~\ref{nmt}, the decoder RNN is a conditional-GRU which is fed by the input context vector, its hidden state and the feedback (\textit{i.e.} the previous generated symbol).
Since we now have two outputs, we need to define what kind of feedback is more suitable for the Factored NMT system.
Several solutions are possible.

The first assumption we made is highly dependent on the design of the considered factors, \ie\ the lemmas are the most informative factors among all.
Then, we tried using only the output lemma embedding as feedback (see equation~\ref{eq:lemma_emb}). 
\begin{equation} \label{eq:lemma_emb}
  \mathrm{Lemma~feedback}: \E_{L}[y_{j}]
\end{equation}
where $\E_{L}$ is the target language lemma lookup table and $\E_{L}[y_{j}]$ is the embedding of the lemma used to generate the output word $y_i$.

\setlength{\belowdisplayskip}{5pt} \setlength{\belowdisplayshortskip}{5pt}
\setlength{\abovedisplayskip}{5pt} \setlength{\abovedisplayshortskip}{5pt}

Another straightforward operation is to sum the embeddings of the previous lemma with the embedding of the previous factors, as described in equation~\ref{eq:sum_emb}.
\begin{equation} \label{eq:sum_emb}
%  \mathrm{Sum~feedback} = (\E_{L}[y_{j}] + \E_{F}[y_{j}]) \cdot \W
  \mathrm{Sum~feedback}: \E_{L}[y_{j}] + \E_{F}[y_{j}]
\end{equation}
where $y_i$ is the target output word, and $\E_{L}[y_i]$ and $\E_{F}[y_i]$ are its corresponding lemma and factors embeddings.
While this could seem unnatural, by doing this, we hope to obtain a joint vector representation of both the lemma and the factors.

Finally, we investigated whether the neural network can learn a better combination of the lemmas and factors embeddings using a linear (eq.~\ref{eq:linear_emb}) or non-linear (eq.~\ref{eq:tanh_emb}) operation instead of a simple sum. 
\begin{eqnarray} 
  \mathrm{Linear~feedback:} &&\E_{L}[y_{j}] \cdot \W_{L} + \E_{F}[y_{j}] \cdot \W_{F} \label{eq:linear_emb}\\
  \mathrm{Tanh~feedback:} && \tanh \left( \E_{L}[y_{j}] \cdot \W_{L} + \E_{F}[y_{j}] \cdot \W_{F} \right) \label{eq:tanh_emb}
\end{eqnarray}
where $\W_L$ and $\W_F$ are the parameters to be learned. % with standard stochastic gradient descent. %SGD.

%The sum of the embeddings gives better score than the linear and tanh transformation but less number of OOV words.
%\todo{give an explanation}
 \begin{table}[!htbp]
\begin{center}
\begin{tabular}{ l|l|l|c|c|c } 
 \hline
  &&\multicolumn{3}{|c|}{\textbf {\%BLEU}} & \\
 \textbf {Model} & \textbf{Feedback} & \textbf{word} & \textbf{lemma} & \textbf{factors} & \textbf {\#OOV} \\
  \hline
NMT  & -		& 34.88 & - 	& - 	& 1775 \\
FNMT & Lemma	& 34.80 & 37.78 & 42.72 & 784 \\ % old model 34.53 in words & 37.46 in lemmas & 42.59 in factors & 787 OOVs
FNMT & Sum    	& 34.48 & 37.14 & 44.46 & 815 \\
FNMT & Linear  	& 34.42 & 37.27 & 44.03 & 868 \\ 
FNMT & Tanh    	& 34.58 & 37.28 & 43.96 & 757 \\ 
 \hline
\end{tabular}
\end{center}
\caption{\label{ta:factors_comb_res}Performance in terms of \%BLEU computed on word, lemma and factors when using different output embedding combinations as feedback. }
\end{table}
\vspace{-8pt}
%Lemma output is the best for this model, factors output the worst. OOvs are reduced but tanh feedback has less.
%I think the lemma feedback has more info of the word an get better translation. 
%We can write that we can explore the concatenation of the 2 embeddings of deeper feedback to learn better the combination.
Table~\ref{ta:factors_comb_res} presents the results obtained with systems integrating the different output embedding combinations as feedback.
%We can see that all systems perform similarly regarding BLEU score on words with a slightly better result for the non-linear combination.
We can see that all systems perform similarly regarding BLEU score on words with a better result for the lemma feedback. %non-linear combination.
As expected, when using only lemma as feedback, the system better estimates the lemmas probabilities, as a consequence, there is a significant reduction of the performance on factors.
The comparison between the lemma \%BLEU (fourth column of Table~\ref{ta:factors_comb_res}) and the number of OOVs (sixth column) shows a correlation between those two values, except when using non-linear combination which has the lowest value of OOVs. %only lemma as feedback. 
%The lower the BLEU is for lemmas, the higher the number of OOV.
This tends to prove that modeling the lemmas better is important to reduce the OOV rate (confirming our assumption that lemmas are more informative) but not sufficient.
In the future we would like to explore the combination of the two embeddings using its concatenation to see if we can get better results.
% merc- I do not understand the line below
%When the factors are worse modeled, the OOV rate increase, which means that the model prefers to generate UNK token with wrong factors instead of another lemma.

%%%%%%%%%%%%%%%%%%%%%%% Dependency model %%%%%%%%%%%%%%%%%%%%%%%%%%%%%%%%%%%%%%%%

\subsubsection{Dependency model}

\begin{tabular}{p{11cm}@{\hskip 25pt} p{3.5cm}}
    \vspace{10pt} 
    One observation that can be made is that while generating factors could seem easier due to the small number of the possible outputs (only 142), the BLEU score is not as high as what we could expect.
However, one could argue that generating a sequence of factors in French from a sequence of English words is not an easy task.
In order to help the factors prediction, we contextualized the corresponding output with the lemma being generated.
This creates a dependency between the lemma output and the factors output. 
The dependency has been implemented by including a transformer (see Figure~\ref{fi:dep_model}) which projects the lemma embeddings into the hidden layer used to generate factors. 
The results by applying those two techniques are presented in Table~\ref{ta:factors_improvement}.& 
    \vspace{0pt}
    \includegraphics[scale=0.3]{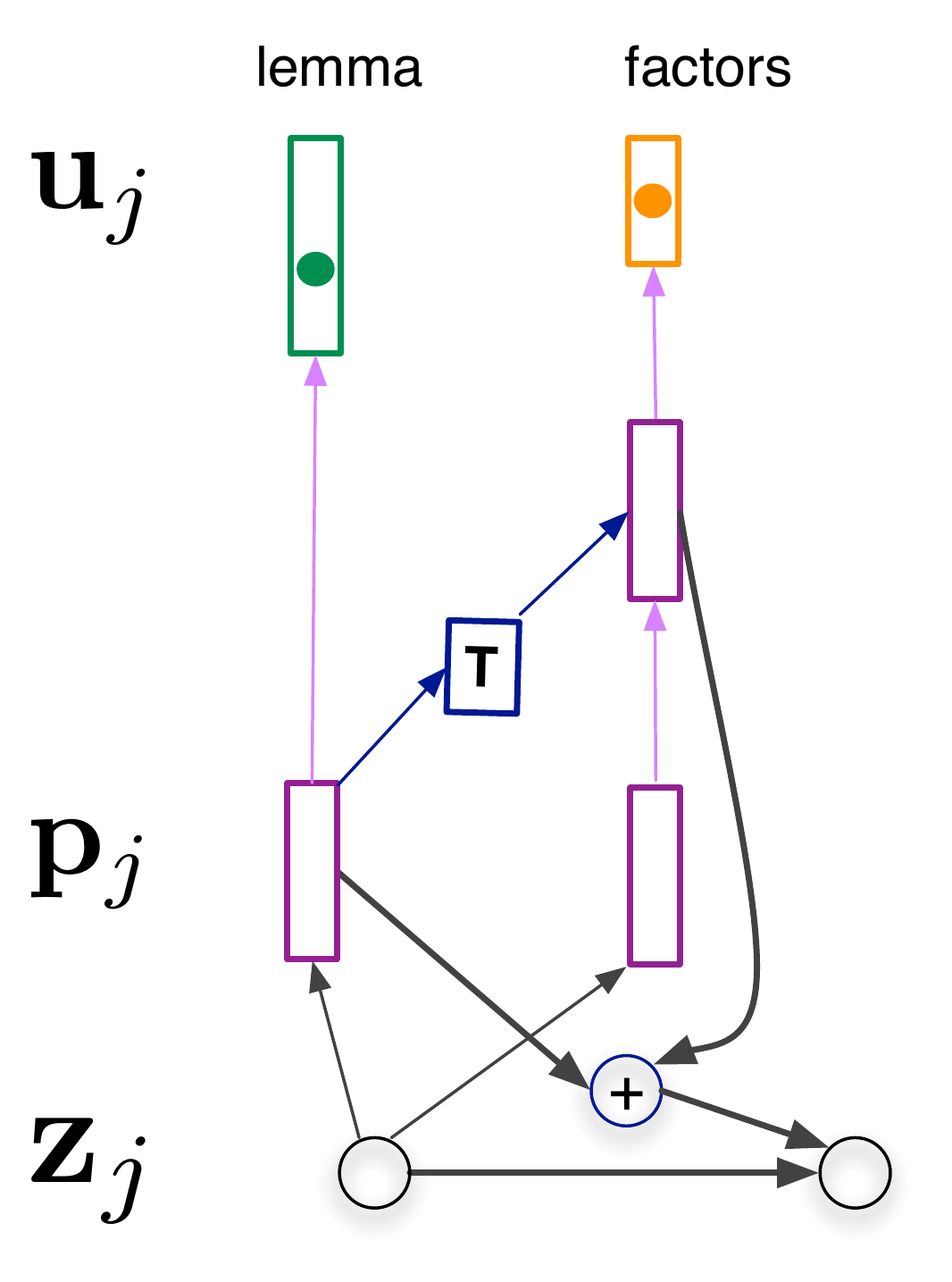}
    \captionof{figure}{\label{fi:dep_model} Dependency model}
\end{tabular}

\begin{table}[!htbp]
\begin{center}
\begin{tabular}{ l|l|l|c|c|l } 
 \hline
  & & \multicolumn{3}{|c|}{\textbf {\%BLEU}}\\
 \textbf {Model} & \textbf{Feedback} & \textbf{word} & \textbf{lemma} & \textbf{factors} & \textbf {\#OOV}\\
  \hline
 NMT & - & 34.88 & - & - & 1775\\
 \hline
% FNMT & Lemma & 34.53 & 37.46 & 42.59 & 787\\ % Lemma feedback 
 FNMT with dependency & Lemma & 34.45 & 37.45 & 42.15 & 770\\%Lemma feedback 
 % FNMT cost weighting & Lemma & 34.56 & 37.50 & 42.53 & 898\\% Lemma feedback
 \hline
% FNMT & Sum     & 34.48 & 37.14 & 44.46 & 815 \\
 FNMT with dependency & Sum & 34.65 & 37.34 & 44.35 & 800\\% Sum feedback 
 \hline
% FNMT & Linear  	& 34.42 & 37.27 & 44.03 & 868 \\ 
 FNMT with dependency & Linear  	& 34.25 & 37.02 & 43.57 & 822 \\ 
 \hline
% FNMT &Tanh     	& 34.58 & 37.28 & 43.96 & 757 \\ 
 FNMT with dependency & Tanh & 34.38 & 37.09 & 43.82 & 915\\% Tanh feedback 
 \hline
\end{tabular}
\end{center}
\caption{\label{ta:factors_improvement}Results for dependency model}
\end{table}

In Table~\ref{ta:factors_improvement}, we can observe that the dependency model does not improve the results in terms of \%BLEU score on words from Table~\ref{ta:factors_comb_res} using lemma, linear and tanh feedback.
However, it improves using the sum feedback. 
For the sum feedback dependency model, we see that lemma BLEU output improves with respect to the same model without dependency. By contrast, the factors output obtains lower BLEU. 
This can occur because factors output receives more information from lemma and when the factors cost is back-propagated, the lemma output can improve the learning. 
We can also observe that if we improve lemma output it is more correlated to the word evaluation than if we improve factors output.
Moreover, the number of the OOV are reduced for all the feedback combination excepting tanh feedback, which is not reflected by the automatic score. 
%\rusure{In contrast, the factors generation does not seem to perform better maybe because the model is generating synonyms or other factored forms that do not correspond to the reference and they are penalized by BLEU score}.
%LB: if we can't say more, then we should not put dependency in this paper

%%%%%%%%%%%%%%%%%%%%%%%%%%%%%%%%%%%%%%%%%%%%%%%%%%%%%%%%%%%%%%%%%%%%%%%%%%%%%
%%%%%%%% Explain deeplem?
%%%%%%%% Results reducing layer sizes?
%%%%%%%%%%%%%%%%%%%%%%%%%%%%%%%%%%%%%%%%%%%%%%%%%%%%%%%%%%%%%%%%%%%%%%%%%%%%%
%%%%%%%% Qualitative analysis
%!TEX root = fnmt.tex

\subsubsection{Qualitative analysis}

We have observed some of the translation outputs to better understand in what cases our FNMT system performs better or worse than the NMT system.

\begin{table}[!htbp]
%\small
\begin{center}
\begin{tabular}{ l|l|l| } 
 \hline

 \multirow{4}{*}{1} & Src & set of \color{red}adaptive \color{black}choices that our \color{red}lineage \color{black}made\\
 & Ref & de choix \color{red}adaptés \color{black}établis par notre \color{red}lignée\\
 & NMT & de choix \color{red}UNK \color{black}que notre \color{red}UNK \color{black}a fait\\
 & FNMT & de choix \color{red}adaptatifs \color{black}que notre \color{red}lignée \color{black}a fait\\
\hline
% \multirow{4}{*}{2} & Source & but here 's where it \color{red}gets \color{black}interesting\\
% & Reference & mais voilà où ça \color{red}devient \color{black}intéressant .\\
% & NMT & mais là , c' \color{red}est \color{black}intéressant . \\
% & FNMT & mais voilà où ça \color{red}devient \color{black}intéressant . \\
% \hline 
%    … ID	Segment 1083, Document "fakedoc" [ΔBLEU=-0.62]
 \multirow{4}{*}{2} & Src	& here 's the \color{red}updated \color{black}version of this entry\\
 & Ref & voici la version \color{red}actualisée \color{black}de cette entrée .\\
 & NMT & voici la version \color{red}mise à jour \color{black}de cette entrée .\\%[0.38]
 & FNMT &	voici la version \color{red}actualisée \color{black}de cette entrée .\\% [1.00]
 \hline
 % ID	Segment 1300, Document "fakedoc" [ΔBLEU=-0.72]
%Source	yes there it is
%Reference oui , le voilà .
% NMT	oui . voilà . [0.28]
% FNMT	oui , le voilà . [1.00]
%ID	Segment 1979, Document "fakedoc" [ΔBLEU=-0.67]
%Source	want to feed the world
%Reference vous voulez nourrir le monde ?
%NMT voulons nourrir le monde . [0.33]
%FNMT vous voulez nourrir le monde ? [1.00]
%%%%%%%%%%%%%%%%% Bad translation %%%%%%%%%%%%%%%%%%%%%%%%%%%%%%
%ID	Segment 2003, Document "fakedoc" [ΔBLEU=0.29]
 \multirow{4}{*}{3} & Src & i \color{red}could \color{black}draw i \color{red}could \color{black}paint\\
 & Ref & je \color{red}pouvais \color{black}dessiner . je \color{red}pouvais \color{black}peindre .\\
 & NMT & je \color{red}pouvais \color{black}dessiner . je \color{red}pouvais \color{black}peindre . \\%[1.00]
& FNMT & je \color{red}pourrais \color{black}dessiner . je \color{red}pouvais \color{black}peindre . \\%[0.71]
  \hline
 %%%%Synonym
%ID	Segment 930, Document "fakedoc" [ΔBLEU=0.46]
\multirow{4}{*}{4} & Src	& \color{red}and \color{black}it 's a very \color{red}easy \color{black}question\\
& Ref & c' est une question très \color{red}simple \color{black}.\\
& NMT & c' est une question très \color{red}simple \color{black}. \\%[1.00]
& FNMT & \color{red}et \color{black}c' est une question très \color{red}facile \color{black}.\\%[0.54]
 \hline
 \end{tabular}
\end{center}
\caption{\label{ta:example_translations}Examples of translations with NMT and Factored NMT. }
\end{table}

\subsubsection*{Translation examples with better BLEU performance}

In the first two examples of Table~\ref{ta:example_translations}, the FNMT system obtains better BLEU score than the NMT system. 

% the words adaptés and lignée are in the French vocabulary but they are predicted as UNK
First example shows when our factored system can generate words when the NMT base system predicts unknown words.
Firstly, the word \textit{lineage} in source sentence is translated as the reference (\textit{ligneé}) by the FNMT system and mapped to UNK by the NMT base system.
Secondly, the word \textit{adaptive} is translated as \textit{adaptatifs} by the FNMT system, the reference translation is \textit{adaptés}, but we can consider the FNMT choice a better translation. NMT system also mapped the word \textit{adaptive} to UNK. 
% The word devient is in the French vocabulary but is not predicted by NMT system
%In the second example FNMT system can generate more precise translation than NMT base system and it is equal to the reference translation.

In the second example, FNMT translation performs as the reference. We are able to generate the new word \textit{actualisée} (\textit{actualiser}+past participle+feminine+singular) that it is not in the shortlist of the NMT system vocabulary. 
This is due, on one hand, because the word \textit{actualisée} appears 40 times in the word vocabulary of the NMT system so it is excluded from the shortlist.
On the other hand, the lemma \textit{actualiser} appears 172 times in the lemmas shortlist so it is included and we are able to generate \textit{actualisée} from the lemma and factors outputs. 
These examples can show the potential of our FNMT system generating new words and reducing unknown words.

\subsubsection*{Translations with lower BLEU performance}

We also have extracted some translations where we have seen a lower BLEU from the FNMT system with respect to the NMT base system (see Table~\ref{ta:example_translations}).

Example 3 shows a problem with the factors output, from the correct lemma \textit{pouvoir}, the FNMT system has generated the word \textit{pourrais} instead of \textit{pouvais}. We can consider both translations as correct but BLEU score penalizes the FNMT translation. 
%FNMT system prefers to translate with incorrect factors before mapping to UNK token. 
%In the example 2, we see that the factored translation is almost correct but it did not translate the word \textit{them}. 
%Moreover, FNMT system used different meaning of the word \textit{make} (\textit{faire/rendre}).
%FNMT system could not translate the word \textit{us} in example 3. 
%This is due to the low diversity of the hypothesis in beam search in this sentence.
%Finally, in the example 3, we saw that the translations are correct for both systems but the reference is equal to the NMT system.

Finally, in the last example, we saw that the translation of the FNMT system is more correct than the NMT system because it translated the word \textit{and} to \textit{et} but in the reference is not included. In addition, FNMT system translated \textit{easy} to a synonym (\textit{facile}) of \textit{simple}.
Consequently, BLEU score penalizes this example in FNMT system being a correct translation.

%Extra tokens, shorter sentences

% multiple words problem
%Some other problems found are that sometimes when we generate the words from the lemmas and its factors, we could have several possibilities of words. This problem is caused because in French we can write the same word in several manners due to the new orthographic rules or the use of a shorter way of writing a word (i.e.: \textit{répét, répétition}). We calculated that we can gain +0.1 of BLEU (34.61) if we are able to solve this ambiguity. We could solve this problem making use of a language model.
\section{Conclusion}
\label{conclusion}
%!TEX root = fnmt.tex

In this paper, we have proposed an NMT architecture which produces a factored representation of the target language words.
Those factors are based on linguistics \textit{a priori} knowledge.
We showed that we are able to train Factored NMT systems with similar performance to word based systems but with the advantage of modeling an almost 6 times bigger word vocabulary with only a slight increase of the computational cost.
A consequence of that is the OOV rate reduction observed with the FNMT system. 
Also, the use of additional linguistic resources allows us to generate new word forms that would not be included in the standard NMT system shortlist.

By reducing the target language vocabulary, we simulated an out-of-domain setup, and we showed that our factored NMT method performs better than the basic NMT system in this case. 
%A further qualitative analysis will be necessary to analyze the problems we found with factors NMT system due to its complexity.
%This approach is promising, we still have margin of improvement, resolving the ambiguity problems with multiple words for the same factors. 

%%%%%%%%%%% Future work %%%%%%%%%%%%%%%%%%%%%%%
% Comparison with more state of the art methods: subwords, large vocabulary paper of groundhog
% Future implementations:
% 1. combination of embeddings for the feedback like their concatenation. 
% 2. dependency from lemmas plus deeper layers. I implemented this model without improvement -> attention_factors_prevlem_deeplem
% 3. dependency plus one factors layer
% 4. more types of weighting in beam search (closer value to 1) and cost in training
% 5. include restriction of valid factors for each lemma in beam search -> multitask
As future work, we would like to include linguistic features at the input.
It is known that this can be helpful for NMT \cite{SennrichH16}.
Extending the approach with input factors could make the target language factors generation simpler. 
This will be investigated in the future.
%Furthermore, we can add the possibility of having more number of outputs to compute more factors independently instead of using their concatenation.
The proposed Factored NMT method could even show better performance if applied on highly inflected languages like German, Arabic, Czech, Russian or Hindi on the target side. 
%Also, we would like to try the method in other corpora with less coverage.

%\section*{Acknowledgements}
%WILL BE ADDED FOR FINAL VERSION
%TODO include M2CR project

\bibliographystyle{acl}
\bibliography{fnmt}

\end{document}